\documentclass[10pt,twocolumn,letterpaper]{article}
\usepackage{wacv}
\usepackage{times}
\usepackage{epsfig}
\usepackage{graphicx}
\usepackage{amsmath}
\usepackage{amssymb}
\usepackage{subcaption}

\wacvfinalcopy % *** Uncomment this line for the final submission

\def\wacvPaperID{653} % *** Enter the wacv Paper ID here

\ifwacvfinal\fi
\begin{document}

%%%%%%%%% TITLE
\title{Euclidean Invariant Recognition of 2D Shapes Using Histograms of Magnitudes of Local Fourier-Mellin Descriptors}

\date{}

%\author[1, 2]{Xinhua Zhang}
%\author[1]{Lance R. Williams}

%\affil[1]{Department of Computer Science, University of New Mexico}
%\affil[2]{New Mexico Consortium}

% % Authors at the same institution
% \author{Xinhua Zhang \hspace{2cm} Lance R. Williams \\
% Department of Computer Science\\
% University of New Mexico\\
% {\tt\small xinhua@unm.edu} \hspace{2cm} {\tt\small williams@cs.unm.edu}
% }
% Authors at different institutions
%\author{Xinhua Zhang \\
%New Mexico Consortium\\
% {\tt\small xinhua@unm.edu}
% \and
% Lance R. Williams \\
%Department of Computer Science\\
%University of New Mexico\\\

\author{Xinhua Zhang and Lance R. Williams \\ Department of Computer Science \\ University of New Mexico}

% \author{Xinhua Zhang \\
% \affiliation{Department of Computer Science, University of New Mexic}
% \affiliation{New Mexico Consortium}
%\author[1,2]{Xinhua Zhang\thanks{xinhua@unm.edu}}
%\author[1]{Lance R. Williams\thanks{williams@cs.unm.edu}}

\maketitle
\ifwacvfinal\thispagestyle{empty}\fi

%%%%%%%%% ABSTRACT
\begin{abstract}
 Because the magnitude of inner products with its basis functions are invariant to rotation and scale change, the Fourier-Mellin transform has long been used as a component in Euclidean invariant 2D shape recognition systems. Yet Fourier-Mellin transform magnitudes are only invariant to rotation and scale changes about a known center point, and full Euclidean invariant shape recognition is not possible except when this center point can be consistently and accurately identified. In this paper, we describe a system where a Fourier-Mellin transform is computed at every point in the image. The spatial support of the Fourier-Mellin basis functions is made local by multiplying them with a polynomial envelope. Significantly, the magnitudes of convolutions with these complex filters at isolated points are not (by themselves)  used as features for Euclidean invariant shape recognition because reliable discrimination would require filters with spatial support large enough to fully encompass the shapes. Instead, we rely on the fact that normalized histograms of magnitudes are fully Euclidean invariant. We demonstrate a system based on the VLAD machine learning method that performs Euclidean invariant recognition of 2D shapes and requires an order of magnitude less training data than comparable methods based on convolutional neural networks.
\end{abstract}
%%%%%%%%% BODY TEXT
\section{Introduction}
The problem of building a system for 2D shape recognition that is invariant to rotation, 
translation and scale change has a long history in computer vision.
Existing approaches to Euclidean invariant recognition have primarily fallen into three categories: 
1) learning invariants by using augmented training datasets; 
2) pooling of non-invariant filter responses; and 3) design of mathematical invariants.
The idea underlying the first category of approaches
is to allow a machine learning system to discover natural invariants by training it on a dataset of images of 
different objects augmented with images of the same objects transformed by rotation, translation and scale change.
For this to work, the augmented training dataset must span the space of different objects and completely 
and densely sample the space of Euclidean transformations.
Several examples of work in this category use convolutional neural networks (CNN)
as the machine learning method because these automatically provide a degree of 
translation invariance \cite{Jaderberg2015,Cheng2016,Esteves2017},
which lessens the amount of training data required.
However, the translation invariance of a CNN is limited by the lack of translation invariance in its 
(non-convolutional) fully-connected layers.
Because translation invariance in practice depends in large part on local pooling operations
in these fully-connected layers, 
training on a dataset augmented by translations is still required.
In summary, despite the potential of these approaches for very high recognition rates, 
the actual degree of Euclidean invariance achieved 
depends almost entirely on the number of images in the augmented dataset;
denser sampling of the space of Euclidean transformations yields 
better invariants but is often impractical.

Unlike approaches in the first category (which augment the training data in order to learn invariants)
approaches in the second category (in effect) augment a set of filters.
More specifically, they pool the responses of a bank of non-invariant filters identical up to Euclidean transformation
with the goal of producing a Euclidean invariant net response
\cite{Sohn2012,Sifre2013,Kanazawa2014}. 
For example, rotating a set of oriented bandpass filters and then averaging the resulting feature maps yields approximate rotation invariance
\cite{Sifre2013}.
Similarly, applying a filter to an image at different scales yields approximate scale invariance \cite{Kanazawa2014}. 
As with the data augmentation approaches, the degree of invariance actually achieved depends
on the density and completeness of the sampling of the space of Euclidean transformations.

\begin{figure}[t]
\centering
\includegraphics[width=0.48\textwidth]{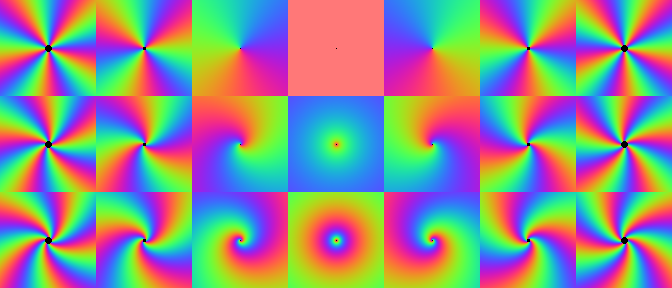}
\caption{Log-polar harmonic gratings.  The phases of complex values are mapped to hue with red indicating positive real. 
Each row is a distinct radial frequency. From top to bottom: 0, 1, 2.  Each column is a distinct angular frequency. 
From left to right: -5, -3, -1, 0, 1, 3, 5.} 
\label{fig:log-polar harmonics}
\end{figure}

As for approaches in the third category, many invariant features are based on harmonic gratings. 
This is because the magnitudes of inner products of images and harmonic gratings
\cite{Simoncelli1996,Mellor2008,Liu2014,Worrall2017} are translation invariant. 
The problem of designing features invariant to the other components of Euclidean transformation
employ coordinate changes that reduce the transformation to translation.
Rotation invariance, for instance, is translation invariance on the angular coordinate in polar coordinates
\cite{Simoncelli1996,Jacovitti2000,Pew-ThianYap2010,Liu2014,Worrall2017}.
Rotation and scale invariance can both be reduced to translation invariance in log-polar coordinates \cite{Ghorbel1994,Gotze2000,Derrode2001,Chi-ManPun2003,Kokkinos2008, Kokkinos2012, Mennesson2014}. 
Interestingly, neuroscience experiments have shown that neurons in areas V2 and V4 of macaque visual cortex are selective for harmonic gratings
in log-polar coordinates \cite{Gallant1993,Gallant1996,Hegde2000}.
See Fig. \ref{fig:log-polar harmonics}. 
Similar polar separable harmonic gratings can be learned by machine learning systems trained on sets of images related by a geometric transformation;
when the images in the sets are related by rotation, the system learns polar harmonic gratings \cite{Memisevic2010};
when the transformations include both rotation and scaling, the system learns log-polar harmonic gratings \cite{Memisevic2011}.

A wide variety of polar separable harmonic grating designs have been used in systems aimed at Euclidean invariant recognition of 2D shapes. 
All of these systems use a harmonic signal in the angular coordinate $\theta$.
Since the angular dimension is periodic, this is the obvious choice.
However, the radial coordinate $r$ is not similarly constrained, and this is where the variety in the different designs resides.
First, the radial coordinate can be logarithmic or non-logarithmic.
The decision to use a logarithmic radial coordinate $\log r$ is the correct one if scale invariance is required.
Second, the function of the radial coordinate might (or might not) include a harmonic signal.
Including a harmonic signal is useful since it increases the discriminative ability of the resulting descriptors 
without compromising scale invariance.
Finally, because the radial dimension is non-periodic, the function of the radial coordinate should
include an envelope to increase locality, and the function used
for this purpose has varied widely.
The specific combination of logarithmic radial variable, harmonic signals in angular and radial variables,
and $\frac{1}{r}$ envelope is embodied in the Fourier-Mellin transform (FMT).
While the $\frac{1}{r}$ envelope provides a weak form of locality while also preserving scale invariance,
it doesn't satisfy the stronger locality constraint defined by \cite{Mellor2008}.
Conversely, while a polynomial envelope satisfies the stronger locality constraint,
it would seem to be incompatible with scale invariance.
Significantly, in this paper we show 
that log-polar harmonic filters with polynomial envelopes can (in fact) be used for scale invariant
recognition if the vector of responses is appropriately normalized.

There are other approaches to achieving translation invariance distinct from harmonic gratings. 
%{\it e.g.} methods based on histograms of gradient directions in neighborhoods of keypoints \cite{lowe}.
Methods based on histograms (or other statistics) of convolutions are inherently translation invariant because spatial information is discarded.
Representative methods assume that inputs have a multimodal distribution and computes a visual vocabulary to model this distribution.
For example,  the Fisher vector method \cite{Perronnin2006} uses Gaussian mixture models as its visual vocabulary,
while the {\it bag of words} \cite{Sivic2003} and {\it vector of locally aggregated descriptor (VLAD)} \cite{Jegou2010}
methods use centers computed by {\it k-means clustering.}

In this paper we propose to use the statistics of locally rotation and scale invariant features to construct globally Euclidean invariant descriptors.
Using feature statistics, {\it e.g.,} of wavelet subbands, rather than the features themselves,
is an approach more commonly used in texture analysis.
We extend this idea and show that it can also be used for Euclidean invariant 2D shape recognition.
Specifically, we apply the VLAD machine learning algorithm to the magnitudes of 
local Fourier-Mellin descriptors---log-polar separable harmonic gratings with polynomial envelopes.
We explore the quality of locality for different choices of envelope and show that descriptors based on polynomial envelopes 
exhibit superior scale invariance and result in better image reconstructions from magnitudes alone. 

\section{Related Work}
Chi {\it et al.} \cite{Chi-ManPun2003} use signatures based on the energy of orthogonal wavelet transforms of
log-polar images as Euclidean invariants.
The technique they describe uses an adaptive row shift-invariant wavelet transform to achieve a degree of
translation invariance in the log-polar domain because (unlike harmonic gratings)
discrete wavelet transforms are not intrinsically translation invariant. 

Esteves {\it et al.} \cite{Esteves2017} train a CNN to recognize objects in log-polar transformed images.
However, a separate CNN is used to learn the object center points before the log-polar transform is applied.
Finally, as with all CNNs, the translation invariance achievable without extra training data is limited,
and the rotation invariant recognition rate would be low if the system were not trained on augmented datasets.

Worrall {\it et al.} \cite{Worrall2017} uses convolutions with harmonic gratings in polar coordinates
as input to a CNN to achieve rotation and translation invariant recognition of 2D shapes.
However, they did not address the question of scale invariance and rotation and translation
invariance are only achieved by training on augmented datasets.
Pew {\it et al.} \cite{Pew-ThianYap2010} use a set of orthogonal basis functions in polar coordinates 
as rotation invariant features for 2D shape recognition.
Like \cite{Worrall2017}, they do not address the issue of scale invariance.

The prior work most similar to our own is by Gotze {\it et al.} \cite{Gotze2000},
who used rotation and translation invariant descriptors based on
magnitudes of inner products with polar separable harmonic gratings with Gaussian envelopes.
To avoid the problem of identification of the center point, these descriptors are computed at
every point in the image using convolution.
The dimensionality of the descriptor computed at each point is relatively low,
and unlike the system we describe, there is no use of the 
statistics of descriptors within extended regions.
Consequently, acceptable recognition accuracy can only be achieved by making the spatial support of
the filters large enough to fully encompass the shape being recognized.
Finally, unlike our system,
which uses polynomial envelopes to achieve scale invariance,
the use of the Gaussian envelope by \cite{Gotze2000}
makes this impossible.

Lehiani {\it et al.} \cite{Lehiani2017} describe a system that uses magnitudes of analytical Fourier-Mellin transform computed 
at keypoints as a Euclidean invariant signature.
Like \cite{Gotze2000}, the dimensionality of the descriptor at each keypoint is relatively low
and acceptable recognition accuracy is only achieved by using Fourier-Mellin basis
functions with large spatial support.

Hoang {\it et al.} \cite{Hoang2010,Hoang2012} describe a Euclidean invariant descriptor which combines the radon transform and magnitudes of a 1D Fourier-Mellin transform. While the basic idea of constructing an invariant descriptor by converting rotation and scaling to translation is 
similar to ours,
%The pattern of 2D Fourier-Mellin function has been used as stimuli to V2 %and V4 neurons and can be learned by certain gated networks as described in %the previous section. 
their descriptor is not spatially localized and our results demonstrate the importance of localization.

%Other work related to our own is described 
Finally, Mellor, Hong and Brady \cite{Mellor2008}
used histograms of responses of polar separable filters with a polynomial envelope for texture recognition.
Our work is different from theirs in three ways: 
1) The filters they use are not
harmonic gratings and are unrelated to Fourier-Mellin basis functions;
2) They use the joint histogram of only two Euclidean invariant features 
while we use VLAD (a non-parametric generalization of Fisher vectors) to represent the joint 
statistics of a large number of Euclidean invariant features; 
3) They only applied their method to Euclidean invariant recognition of textures, while our method is also applied to
Euclidean invariant recognition of 2D shapes.

\section{Euclidean Invariant Descriptors}

An output representation 
is {\it equivariant} with respect to a geometric transformation 
if the operation that produces it commutes with the 
transformation.
For example,
convolution commutes with rotation 
and translation of the input image if the filter is isotropic.
In contrast, an output representation 
is {\it invariant} with respect to a geometric transformation 
if it is unchanged by it.
For example, histograms are unchanged
by rotation and translation of the input image.

\begin{figure}[t]
\centering
\includegraphics[width=0.48\textwidth]{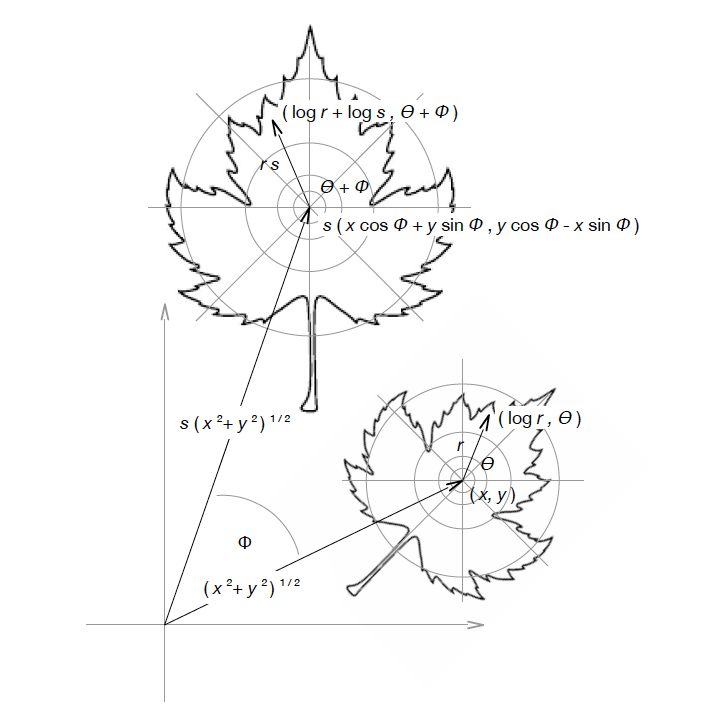}
\caption{An illustration of the relationship between local transforms and global transforms. When a patch is rotated and scaled with with respect to the origin, all points of the patch are rotated and scaled by the same amount with respect to the center of the patch.}
\label{fig:Invariant Leaf}
\end{figure}

\subsection{From Local to Global Invariance}

Consider an image $I$ with a gray value histogram $H$.  A Euclidean
transformation ${\cal T}$ of $I$ about its origin consists of a 
scale change $s$,
rotation by an angle $\phi$, and translation by a vector, $[u,v]^{\rm T}$.
This transformation maps the image and its histogram as follows:
\begin{eqnarray*}
I\left(\left[\begin{array}{c}
x\\
y
\end{array}\right]\right)
& \stackrel{\cal T}{\rightarrow} &
I\left(\left[\begin{array}{c}
s \: (\: x \: \cos \phi + \: y \: \sin \phi) +  u\\
s \: (\: y  \: \cos \phi - \: x \: \cos \phi) + v
\end{array}\right]\right)\\
H(i) & \stackrel{\cal T}{\rightarrow} & s^2 \; H(i).
\end{eqnarray*}
\noindent Because scaling an image by $s$ increases the size of areas by $s^2$,
and because the histogram represents the area possessing a single gray value, 
$H$ is multiplied by $s^2$.
It follows that $H \: / \: ||H||_2$ is invariant with respect to
Euclidean transformation.

It is possible to construct rotation and scale invariant features 
with spatial support greater than one pixel.
A Fourier-Mellin basis function $f_{\omega_r, \omega_\theta}$ is a separable function in log-polar coordinates:
\[
f_{\omega_r, \omega_\theta} = e^{-j \omega_r \log r} e^{-j \omega_\theta \theta} r^{-1}.
\]
%It is the product of a harmonic signal of frequency $\omega_r$ in 
%the log radial coordinate $\log r$,
%a harmonic signal of frequency $\omega_\theta$ in the angular coordinate $\theta$,
%and an envelope that is a polynomial in the radial coordinate $r$.
The magnitude of the inner product of a Fourier-Mellin basis function 
and a circular image patch $p$ is invariant to rotation and scaling 
of the patch because these are translation in log-polar coordinates
and magnitude is not changed by translation:
\[
%| \left<f_{mn}, \: p(\log r, \theta)\right> | = | \left< f_{mn}, \: p(\log r + \log s, \theta + \phi) \right>|
| \left<f_{\omega_r, \omega_\theta}, \: p(\log r, \theta)\right> | = | \left< f_{\omega_r, \omega_\theta}, \: p(\log r \!+\! \log s, \theta \!+\! \phi) \right>|.
\]
\noindent In an image consisting of a set $P$ of overlapping circular patches, a patch $p$ is accessed by specifying its 
{\it global} position in Cartesian coordinates $[x,y]^{\rm T}$. 
Gray values are accessed by specifying their
{\it local} positions within $p$ in log-polar coordinates 
$[\log r, \theta]^{\rm T}$.
Now, consider the effect of the Euclidean transformation ${\cal T}$ on the set of patches and a histogram $H_{\omega_r, \omega_\theta}$ of the magnitudes of inner products of the patches with a Fourier-Mellin basis function $f_{\omega_r,\omega_\theta}$:
{\small
\begin{eqnarray*}
P\left(
\left[\begin{array}{c}
x\\
y\\
\hline
\log r\\
\theta
\end{array}\right]
\right)
& \stackrel{\cal T}{\rightarrow} &
P\left(\left[\begin{array}{c}
s \: (\: x \: \cos \phi + y \: \sin \phi )+  u\\
s \: (\: y  \: \cos \phi - x \: \cos \phi) + v\\
\hline
\log r + \log s\\
\theta + \phi
\end{array}\right]
\right)\\
H_{\omega_r, \omega_\theta}(i) & \stackrel{\cal T}{\rightarrow} & s^2 \; H_{\omega_r, \omega_\theta}(i).
\end{eqnarray*}}
Although the patch centers move, 
the magnitudes of inner products 
with $f_{\omega_r,\omega_\theta}$
are invariant with respect to the local geometric transformation
induced by ${\cal T}$ for all patches
because this is just translation by $[\log s, \phi]^{\rm T}$.
Consequently, magnitude of convolution with $f_{\omega_r,\omega_\theta}$
commutes with the global transformation ${\cal T}$. 
%(consisting of translation by $[u,v]^{\rm T}$,
%rotation by $\phi$ and scaling by $s$)
%for all images $I$.
It follows that 
$|I * f_{\omega_r,\omega_\theta}|$ is 
{\it equivariant} with respect to Euclidean transformation.
Finally, because $|I * f_{\omega_r,\omega_\theta}|$ is equivariant 
with respect to
Euclidean transformation,
$H_{\omega_r, \omega_\theta} \: / \: ||H_{\omega_r, \omega_\theta}||_2$ is 
{\it invariant} with respect to Euclidean transformation.
See Fig. \ref{fig:Invariant Leaf}.

\subsection{Local Fourier-Mellin Transform}
\label{sec:Local Fourier-Mellin Transform}
Given an image $I(r,\theta)$ in polar coordinates, the local Fourier-Mellin transform $LFMT(\omega_r,\omega_\theta,\alpha)$ is defined as 
\begin{equation*}
\label{eq:Local Fourier-Mellin Transform}
%{\textstyle
\frac{1}{2\pi} \int_0^{2\pi}\int_0^{R} I(r,\theta) e^{-j \omega_r \log r} e^{-j \omega_\theta \theta} r^{\alpha} d r d \theta
%}
\end{equation*}
\noindent where $\omega_r$ and $\omega_\theta$ are radial and angular frequencies, $R$ is the range of the transform, and $\alpha$ defines the envelope shape.
Fourier-Mellin transform (FMT) is the case when $\alpha = -1$.
The envelope is scale invariant because the Mellin transform implies $\frac{d r}{r} = \frac{d (sr)}{sr}$.
Therefore, the magnitude of FMT is scale invariant:
\begin{equation*}
	\begin{split}
		&\quad\left|\: \frac{1}{2\pi} {\textstyle \int_0^{2\pi}\int_0^{sR} I \left(r / s,\theta \right) e^{-j \omega_r \log r} e^{-j \omega_\theta \theta} \frac{d r}{r} d \theta } \:\right| \\
		&= \left| \: \frac{1}{2\pi} {\textstyle \int_0^{2\pi}\int_0^{R} I\left(r',\theta\right) e^{-j \omega_r \log r'} e^{-j \omega_\theta \theta} \frac{d r'}{r'} d \theta } \:\right|
	\end{split}
\end{equation*}
\noindent where $r' = \frac{r}{s}$ and $|\cdot|$ is magnitude.
For other values of $\alpha$, there is an extra term $s^{\alpha + 1}$, because of $(sr)^\alpha d (sr)$.
Consequently, the magnitude of the transform is not scale invariant.
Yet the L2 normalized vector comprised of magnitudes at all 
frequencies is.
Several works \cite{Simoncelli1996,Mellor2008,Liu2014,Worrall2017} 
use relative phases instead of magnitudes because relative phases 
are both rotation and contrast invariant. 
We use magnitudes in our work because the number of relative phases is $M^2 \times N^2$, and this is quite large. 
Since the vector of magnitudes can be made contrast invariant
by dividing it by its L2 norm, magnitudes are the better choice.

In order to compute the LFMD, we need to derive a set of convolution kernels by discretely
sampling log-polar harmonic gratings with polynomial envelopes.
Several issues need to be addressed.
First, the kernels should have odd size so that the center of the kernel is pixel-centered and because
$\log 0 = -\infty$, the center of the kernel should be 0. 
Second, because the kernels are constructed by discrete sampling of log-polar harmonic gratings,
the problem of aliasing in the $\theta$ dimension becomes significant
as the angular frequency $\omega_\theta$ increases.
The solution would seem to be to modify the envelope function by zeroing the kernel within
a radius large enough to prevent aliasing of a given angular frequency.\footnote{
Interestingly, CNNs trained on rotation augmented datasets seem to discover the same solution, 
{\it i.e.,} functions resembling polar harmonic gratings with holes in their centers
\cite{Memisevic2010,Memisevic2011}.}
Accordingly, we define a log-polar harmonic grating with polynomial envelope
and a hole in the center as follows
\begin{equation*}
{\small
h_{\omega_r,\omega_\theta,\alpha}(x,y) =
}
\end{equation*}\begin{equation*}
{\small
\label{eq:Local Fourier-Mellin filter}
\left\{
			\begin{array}{ll}
				0 &\!\!{\rm if} \sqrt{x^2 + y^2} \leq \sigma \omega_\theta \\ 
	\frac{1}{2\pi}(\sqrt{x^2 + y^2})^{(\alpha - j \omega_r)} e^{-j \omega_\theta {\rm atan} \left( \frac{x}{y}\right)} &\!\!\mathrm{otherwise.}
			\end{array}
 \right.
}
\end{equation*}
\noindent where the $\exp (- j \omega_r \log r)$ has been replaced 
by $r^{-j \omega_r}$.
Experiments show that $\sigma = \frac{2}{\pi}$ 
gives the best performance.
Finally, given an image $I$, a local Fourier-Mellin descriptor $LFMD_\alpha$ at pixel $(i,j)$ is defined as
\begin{equation*}
\label{eq:convolutional Local Fourier-Mellin Descriptor}
n\left(\left[|\{I * h_{\omega_{r_1}\!,\omega_{\theta_1}\!,\alpha}\}(i,j)|, \dots, |\{I * h_{\omega_{r_M}\!,\omega_{\theta_N}\!,\alpha}\}(i,j)| \right]^{\rm{T}}\right)
\end{equation*}
\noindent where $n(\vec{x}) = \vec{x} \; / \; \| \vec{x} \|_2$.
\subsection{Envelope Locality and Image Reconstruction}
\label{sec:Locality of Polynomial Envelop and Image Reconstruction}
\begin{figure}[t]
	\centering
	\begin{subfigure}[h]{0.2\textwidth}
	\includegraphics[width=1\textwidth]{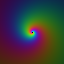}
	\caption{}
	\end{subfigure}
	~
	\centering
	\begin{subfigure}[h]{0.2\textwidth}
	\includegraphics[width=1\textwidth]{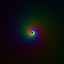}
	\caption{}
	\end{subfigure}
	~
	\centering
	\begin{subfigure}[h]{0.2\textwidth}
	\includegraphics[width=1\textwidth]{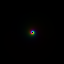}
	\caption{}
	\end{subfigure}
	~
	\centering
	\begin{subfigure}[h]{0.2\textwidth}
	\includegraphics[width=1\textwidth]{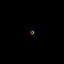}
	\caption{}
	\end{subfigure}
	\caption{Effects of $\alpha$ on Local Fourier-Mellin filters. Filter size is $64\times 64$. (a) $\alpha$ = -0.5  (b) -1 (c) -2 (d) -3.}
	\label{fig:effects of alpha}
\end{figure}

Different envelopes are used for different purposes. 
The $\alpha = -1$ envelope of the FMT is best for scale invariance. 
Analytical Fourier-Mellin transform (AFMT)
\cite{Ghorbel1994,Derrode2001} is the case when  
% $\alpha = \beta - 1$, where $\beta > 0$, thus 
$\alpha > -1$. 
% (the $\beta$ term is needed to ensure the convergence of 
% the Mellin integral).
According to Mellor, Hong and Brady \cite{Mellor2008}, 
a filter $g(r)$ is localized if
\begin{equation*}
\label{eq:locality constrain}
%{\textstyle
\int_0^R 2\pi r g(r)^2 d r > \int_R^\infty 2\pi r g(r)^2 d r
%}
\end{equation*}
\noindent for all $R > 0$.
Envelopes satisfy this constraint when $\alpha < -1$. 
Consequently, the envelopes used in FMT and AFMT
are not localized.
The effect of different values of $\alpha$ on the 
appearance of the filters is shown in Fig. \ref{fig:effects of alpha}.
Because the pros and cons
of the choice of envelope shape are unclear,
we evaluate the effect of envelope shape empirically by
gauging performance
on two problems:
image classification (Section \ref{sec:Image Classification})
and image reconstruction (Section \ref{sec:Image Reconstruction}).

For invertible transforms, such as the polar harmonic transform \cite{Pew-ThianYap2010} or the AFMT \cite{Derrode2001},
the original image can be recovered simply by applying the inverse transform.\footnote{Of course if the basis functions
of the transform are localized, 
and the image size is larger than the spatial support of these functions, 
then it will be unable to recover the entire image.}
However, there is no way to recover the original image from the transform coefficient magnitudes alone.
Since these are the actual Euclidean invariants that will be used 
for 2D shape recognition,
the question of whether or not they are sufficient by themselves for reconstructing 
the original image is not purely academic.
It has been previously shown that images can sometimes be reconstructed from the magnitudes of values
resulting by convolution with complex-valued filters \cite{Shams2002}. 
Although there is a sign ambiguity in the reconstruction, and this is unavoidable since magnitudes discard sign, 
the qualities of the reconstructions are otherwise good.
The results described in \cite{Shams2002}
uses the magnitudes of complex Gabor wavelets.
Because they are the products of harmonic gratings in Cartesian coordinates and Gaussian envelopes,
these filters possess local translation invariance.
In this paper, we show that images can be reconstructed from the 
magnitudes of LFMDs, the products of harmonic gratings in log-polar coordinates and polynomial envelopes,
and which possess local rotation and scale invariance.

Given an image $I$, a reconstructed image $\tilde{I}$, and a local Fourier-Mellin filter $h_{\omega_r,\omega_\theta, \alpha}$, the
reconstruction error $E$ is defined as follows
\begin{equation*}
\label{eq: reconstruction error}
	\begin{split}
	 \sum_{\omega_{r}}\sum_{\omega_{\theta}}
{\left( \left| \frac{I * h_{\omega_r, \omega_\theta, \alpha} \phantom{\tilde I}}{\||I * h_{\omega_r, \omega_\theta, \alpha}|\|_2}\right|^2 - \:\:
\left|\frac{\tilde{I} * h_{\omega_r, \omega_\theta, \alpha}}{\||\tilde{I} * h_{\omega_r, \omega_\theta, \alpha}|\|_2} \right|^2\right)^2}.
	\end{split}
\end{equation*}
\noindent At every position, the convolution is divided by the L2 norm of magnitudes across frequencies. 
The reconstruction $\tilde{I}$ is initialized randomly and updated by gradient descent where
$\Delta\tilde{I}(x,y) = \frac{d E}{d \tilde{I}(x,y)}$. 
The normalization is necessary for LFMDs to be scale and contrast invariant.
Although the definition of error in \cite{Shams2002} does not have the normalization term, it is otherwise the same as the one we used in
the present work.
\subsection{LFMD-VLAD}
Besides standard VLAD \cite{Jegou2010}, we also use the power-law normalization.
More specifically, given a vector $\vec{v} = [v_1,\dots, v_N]^{\rm T}$, 
a normalized vector $\vec{v}^\prime$ equals $[v'_1,\dots,v'_N]^{\rm T}$ where $\vec{v}^\prime_i = |v_i|^\beta \times \mathrm{sign}(v_i)$ \cite{Jegou2012a}. 
The scale, translation and contrast invariant descriptor, LFMD-VLAD, is computed using the following procedure:
\begin{enumerate}
	\item Compute a $MN$ dimensional LFMD at every point.
	\item Compute $K$ centers, $\vec{c}_k$, of LFMDs using $k$-means clustering.
	\item Compute $\vec{v}_k \! = \! {\textstyle \sum_{\{\vec{x} \: | \: \vec{n}(\vec{x}) = \vec{c}_k\}} {\vec{x} - \vec{c}_k}}$ where $\vec{n}(\vec{x})$ is the nearest neighbor of $\vec{x}$. % relation.
	\item Construct a $K \times MN$ dimensional vector $\vec{V}$ by concatenating the $\vec{v}_k$.
	\item Apply $\beta = 0.5$ power law normalization followed by L2 normalization to $\vec{V}$.
\end{enumerate}
\subsection{Scale Invariance and Object Background Ratio}
\label{sec:Object-background Ratio}
When an image of an object is scaled, the ratio of the area of the object and the area of the background changes.
This affects the LFMD magnitude histogram:
\begin{eqnarray*}
H(0) & \rightarrow & H(0) + {\textstyle \sum_{i > 0} \left[ H(i) - s^2 H(i) \right]}\\
H(i) & \rightarrow & s^2 H(i).
\end{eqnarray*}
\noindent where $H$ is the histogram of an image of an isolated object against a uniformly zero valued background.
In a case such as the above, the problem can be solved by thresholding. 
However, the problem is more complex in natural images, and other techniques (beyond the scope of the present work) are needed.

\section{Image Classification}
\label{sec:Image Classification}
In this section, we gauge the degree to which the LFMD-VLAD system 
achieves Euclidean invariant recognition by testing it
in three different problem domains:
1) the MNIST database \cite{Lecun1998} for handwritten digit recognition;
2) the Swedish leaves database \cite{Soderkvist2001} for 2D shape recognition;
and 3) the KTH-TIPS database \cite{Fritz2004} for texture recognition.

We apply rotation and scaling to the MNIST and Swedish leaves databases. 
All transforms use bicubic interpolation.
Our classifier is trained on original images and tested on transformed images.
We did not augment the set of training images in anyway, {\it e.g.,} by inclusion of additional images related to the training set 
by translation, rotation and scaling.
We demonstrate that LFMD-VLAD does not need to be trained on augmented datasets to achieve Euclidean invariant recognition.
In all classification experiments, $\omega_r \in \{0,1,2\}$ and $\omega_\theta \in \{-5,-4, \dots,  5\}$.
The dimensionality of the descriptor, $MN$, is therefore $3 \times 11 = 33$. 
There  are $K = 64$ centers computed by the $k$-means clustering algorithm so the dimension of
the LFMD-VLAD system is $MN \times K = 33 \times 64 = 2112$.
For color images, the same procedure is applied to the three channels independently, 
so the descriptor dimensionality is three times that of gray scale images. 
We use the linear SVM described in \cite{Fan2008} with the parameter $C = 1$.  
Several different values of $\alpha$ are evaluated in order to ascertain the effect of using different envelopes.
When $\alpha$ equals 0, the filters are non-localized log-polar harmonic gratings;
when $\alpha$ equals 0.5, the filters implement the analytical FMT described by \cite{Derrode2001};
when $\alpha$ equal to -1, the filters are the basis functions of the classical FMT.
Finally, we will show experimentally that setting $\alpha$ equal to -2 or -3 gives the best 
locality without sacrificing scale invariance.

We use the algorithm described by \cite{Esteves2017} as a benchmark to compare with our own on the MNIST database \cite{Lecun1998}.
Given an image, it is first log-polar transformed into a $256\times256$ image using the center of the image.
Afterwards,
an AlexNet \cite{Krizhevsky2012} with $224 \times 224$ random cropping and random flipping is trained on the log-polar transformed images. The central idea of this approach is to convert rotation and scaling in Cartesian into translation in log-polar, and CNNs can build up translation-invariance by local pooling layer after layer. Regarding the methods proposed in \cite{Esteves2017}, we do not use the same CNNs because they are not better than AlexNet with respect to translation-invariance in theory. We do not use the origin predictor either, because our objective is not to maximize recognition accuracy,
but to illustrate the degree to which the CNN depends on the use of augmented training sets to achieve Euclidean invariance.
\begin{table}[t]
\begin{center}
\caption{MNIST classification error rates ($\%$) using a training set which consists of  only original images. 
For testing sets, {\it O} is original, {\it R} is rotated, {\it S} is scaled and {\it R+S} is rotated and scaled. Names for methods and training sets are in the first column. For training sets, {\it O} is original, {\it L} is log-polar transformed and {\it +} indicates an {\it R+S} augmented training set.}
\label{tab:MNIST}
%\def\arraystretch{1.25}
%\begin{tabular}{l@{\qquad\qquad}|c|@{\quad}c|@{\quad}c|@{\quad}c|@{\quad}c|}
\begin{tabular}{|c|c|c|c|c|c|}
\hline
   & $\alpha$ & O & R & S & R + S \\ \hline
AlexNet (O) & --   & 1.27 & 52.31  & 7.33 & 58.48 \\\hline
AlexNet (O+) & --  & 5.28 & 5.31 & 6.04 & 6.14 \\\hline
AlexNet (L) & --  & \textbf{1.09} & 37.56  & 2.03 & 41.44\\\hline
AlexNet (L+) & --  & 3.87 & 3.75 & 3.91 & 3.72 \\\hline
LFMD-VLAD &  0 & 5.01 & 5.01 & 5.08 & 5.03 \\\hline
LFMD-VLAD & -0.5 & 3.92 & 3.9 & 4.12 & 4.01 \\\hline
LFMD-VLAD & -1 & 2.75 & \textbf{2.78} & \textbf{3.09} & \textbf{3.12}\\\hline
LFMD-VLAD & -2 & 2.46 & 3.02 & 5.98 & 6.7 \\\hline
LFMD-VLAD & -3 & 7.04 & 13.34 & 21.26 & 27.11\\\hline
\end{tabular}
\end{center}
\end{table}
\subsection{MNIST}
The standard MNIST database contains 60,000 gray scale images of digits of size $28 \times 28$.\footnote{We omitted the digit `9' 
since the purpose of our research is Euclidean invariant recognition and the `9' cannot be distinguished from `6' when rotation is permitted.}
To create the augmented datasets used for testing (and training also in the case of the CNN), each image was resized to $48\times48$,
geometrically transformed in one of the following ways: 
\begin{enumerate}
\item Original: No transformation.
\item Rotation: A random rotation between $0$ and $2 \pi$
\item Scaling:  A random scaling between 0.5 and 1.5
\item Rotation and scaling: The combination of the rotation and scaling transformations defined above.
\end{enumerate}
\noindent After the geometric transformation was applied, the image was padded to size $96\times96$.
In this way, four sets of images were created. We do not augment the dataset in like \cite{Esteves2017} because we want to show that the proposed descriptor can handle transformations in a wider range. Also, this can decrease performance for CNNs more than using a smaller range.
For every transform, the training set has 9,000 images (1,000 per category) images, 
and the testing set has 45,000 images (5,000 per category) images.
Descriptors used by $k$-means are extracted from the first 1,000 images with stride 4, 
but those used for constructing LFMD-VLAD for one image are extracted with stride 1.

\begin{figure}[t]
\centering
\includegraphics[width=0.52\textwidth]{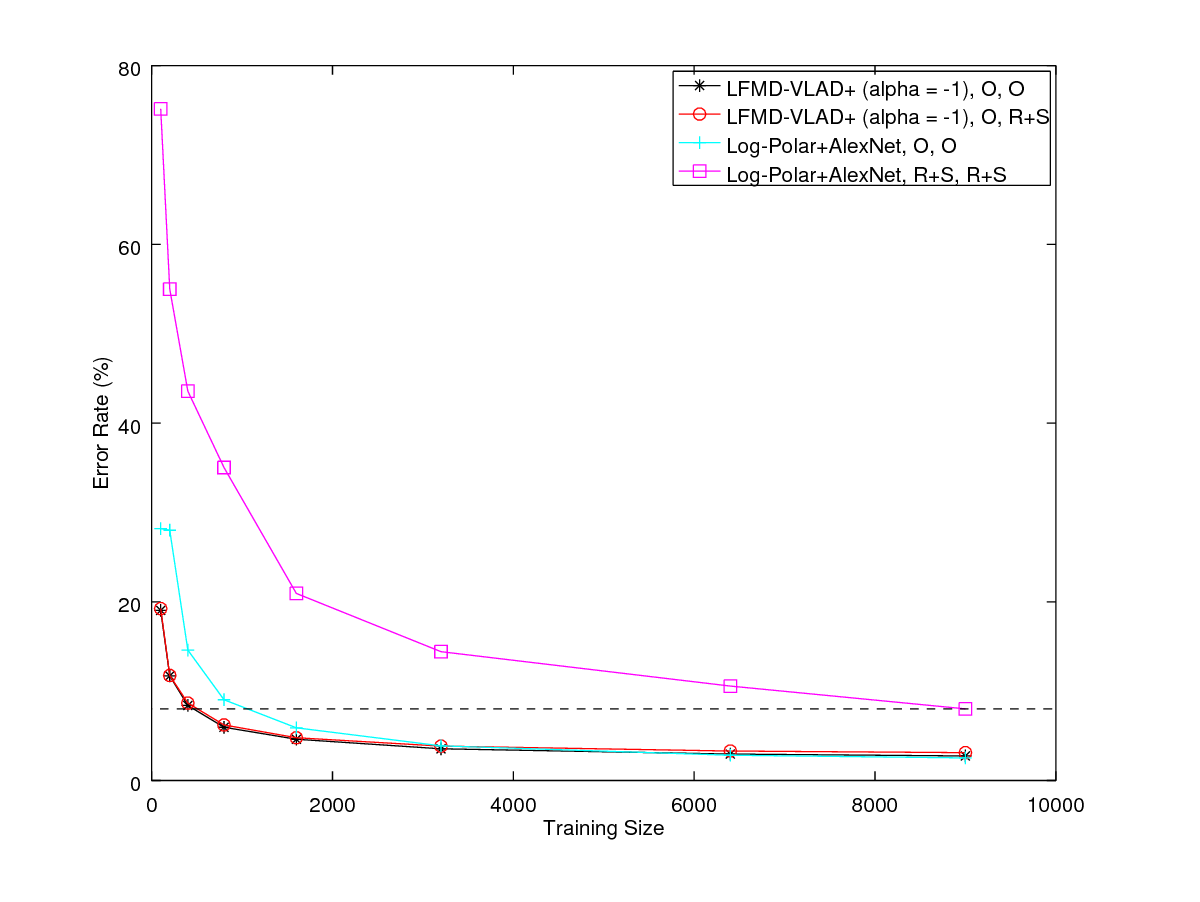}
\caption{Effects of training set size. The format of the legend is \textit{algorithm, training set, testing set}, where {\it O} is original and {\it R+S} is rotated and scaled.
The dash line marks the best error rate achieved by the CNN when trained on the augmented dataset.}
\label{fig:MNIST Trend}
\end{figure}

The results are shown in Table \ref{tab:MNIST}.
AlexNet fails when used with log-polar filters when trained on a non-augmented dataset
because the CNN does not provide enough translation invariance.
The fact that it is more scale invariant than rotation invariant suggests that it is unable to learn the periodic nature of the $\theta$ dimension.
Although it is able to learn rotation and scale invariance when trained on an augmented dataset, 
this requires approx. 10 times more training data than is required by the 
LFMD-VLAD approach (Fig \ref{fig:MNIST Trend}).

\subsection{Swedish leaves}
The Swedish leaves database consists of 15 categories and 75 images per category and is primarily used for 2D shape recognition.
In our experiments, 375 images (25 per category) are randomly chosen to serve as the training dataset and the remaining 750 images serve as the testing dataset.
The same geometric transforms that were used in the MNIST dataset experiment were applied
to construct the augmented testing dataset except that the scale factor is between 0.5 and 1.
Afterwards, images are padded to size $256\times256$.
The process of creating training and testing datasets is repeated five times and the reported classification accuracies are the average values achieved.
The descriptors used for $k$-means clustering are extracted from all training images with stride 4.

Although the original leaf images appear to have a white background, the background is not uniform.
To remedy this, the images are preprocessed as follows:
1) gray values are rescaled to a range between 0 and 255 and then inverted;
2) gray values less than the average for the image are set to zero.
This results in leaf images with uniformly black backgrounds.

Data augmentation sometimes makes AlexNet's performance worse. Since the R+S augmentation only transforms every single image once, it always reduces the number of similar images between training set and testing set. The fact that there are only 375 training images makes this phenomenon more obvious than in the MNIST experiment. We observe that the optimum value of $\alpha$ is different than it was for the MNIST experiment, 
with more localized filters ($\alpha = -3$) yielding increased accuracy. Both shape recognition algorithms \cite{Ke2014, Qi2014} use rotation-invariance to facilitate shape recognition. However, 
neither shows how much invariance is actually achieved and LFMD-VLAD's recognition accuracy is competitive with both on the leaf dataset.
\begin{table}[t]
\begin{center}
\caption{Swedish leaves classification accuracy ($\%$) using a training set which consists of only original images. 
For testing sets, {\it O} is original, {\it R} is rotated, {\it S} is scaled and {\it R+S} is rotated and scaled. Names for methods and training sets are in the first column. For training sets, {\it O} is original, {\it L} is log-polar transformed and {\it +} indicates an
{\it R+S} augmented training set.}
\label{tab:leaves}
%\begin{tabular}{|@{\quad}cl@{\quad}c@{\quad}c@{\quad}c@{\quad}c}
\begin{tabular}{|c|c|c|c|c|c|}
\hline
&  $\alpha$ & O & R & S & R + S \\ \hline
\cite{Ke2014} & --  & 95.33  & --  & -- & -- \\ \hline
 \cite{Qi2014} &  -- & \textbf{99.38}  & --  & -- & -- \\ \hline
AlexNet (O) & --  & 98.13 & 28.40 &  75.67  & 24.72  \\\hline
AlexNet (O+) &  -- & 92.51 & 40.75 & 89.76 &  39.25 \\\hline
AlexNet (L) & --  & 75.15 & 74.75 & 79.47 & 82.37  \\\hline
AlexNet (L+) &  -- &  73.76 & 78.21 & 80.11 & 81.83  \\\hline
LFMD-VLAD  & 0 & 91.46 & 84.66 & 79.90 & 78.2  \\\hline
LFMD-VLAD & -0.5 & 92.8 & 85.84 & 82.04 & 80.8  \\\hline
LFMD-VLAD & -1 & 94.18  & 88.6  & 84.28  & 82.52 \\\hline
LFMD-VLAD & -2 & 98.14 & 96.6 & 78.96 & 78.48\\\hline
LFMD-VLAD & -3 & 99.14 & \textbf{98.40} & \textbf{95.92} & \textbf{95.84}\\\hline
\end{tabular}
\end{center}
\end{table}
\subsection{KTH-TIPS}
KTH-TIPS contains 10 categories and 81 texture images per category. 
Images from one category vary in scale, orientation and contrast.
For this database, we do not need to apply any geometric transformations because they are already included in the dataset.
For cross validation purposes, training datasets are created in different sizes: 50 (5 per category), 200 (20 per category) and 400 (40 per category). 
The remaining images are used to define the testing dataset. 
Images are chosen at random and this process is repeated five times.
The classification accuracy is the average of these five repetitions.
Because textures have stationary spatial statistics, the issue with uniform background is not relevant for this dataset.
Consequently, no thresholding of the LFMD values was used.

The algorithm proposed in \cite{Sifre2013} is state-of-the-art for the KTHTIPS dataset. It outperforms LFMD-VLAD in the 20 per category and 40 per category cases. However, LFMD-VLAD has the best performance in the 5 per category case. Increasing the number of training images does not improve LFMD-VLAD's performance as much it does in the 
$\alpha = -2$ or $\alpha = -3$ cases. 
This suggests a weak relationship between Euclidean-invariance and texture recognition rate.
\begin{table}[t]
\begin{center}
\caption{KTHTIPS classification accuracy ($\%$). Names for methods and training sets are in the first column. {\it O} stands for trained on original images and {\it L} stands for trained on log-polar transformed images. }
\label{tab:KTHTIPS}
\begin{tabular}{|c|c|c|c|c|c|}
%\begin{tabular}{l@{\quad\quad\quad}c@{\quad}c@{\quad}c@{\quad}c}
\hline
& $\alpha$ & 5 & 20 & 40 \\
\hline
\cite{Sifre2013} & -- & 84.3 & \textbf{98.3 }  & \textbf{99.4 } \\\hline
AlexNet (O) & -- & 79.29 & 86.98 & 93.42 \\\hline
AlexNet (L) & -- & 57.92 &  73.02 & 78.83 \\\hline
LFMD-VLAD & 0 & 66.1 &  87.8 & 88.8  \\\hline
LFMD-VLAD & -0.5 & 66.7  & 86.0  & 89.3 \\\hline
LFMD-VLAD &  -1 & 71.5  & 88.3 & 90.3 \\\hline
LFMD-VLAD &  -2 & \textbf{85.5}  & 95.2  & 95.4 \\\hline
LFMD-VLAD &  -3 & 85.3  & 94.8  & 95.7 \\\hline
\end{tabular}
\end{center}
\end{table}

\section{Image Reconstruction}
\label{sec:Image Reconstruction}
\begin{figure}[t]
	\centering
	\begin{subfigure}[h]{0.14\textwidth}
	\includegraphics[width=1\textwidth]{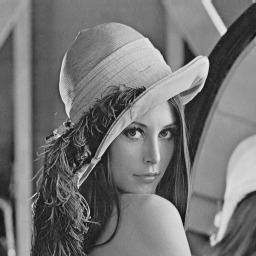}
	\caption{}
	\label{fig:Lena original}
	\end{subfigure}
	~
	\centering
	\begin{subfigure}[h]{0.14\textwidth}
	\includegraphics[width=1\textwidth]{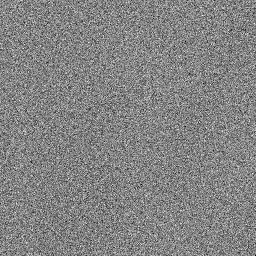}
	\caption{}
	\label{fig:Lena -1}
	\end{subfigure}
	~
	\centering
	\begin{subfigure}[h]{0.14\textwidth}
	\includegraphics[width=1\textwidth]{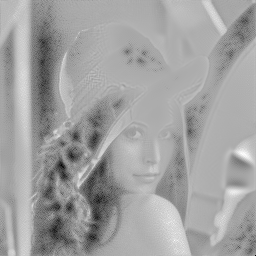}
	\caption{}
	\label{fig:Lena -2}
	\end{subfigure}
	\\
	\centering
	\begin{subfigure}[h]{0.14\textwidth}
	\includegraphics[width=1\textwidth]{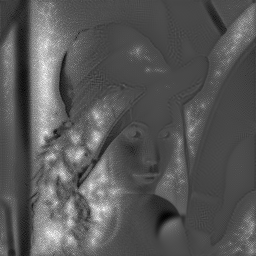}
	\caption{}
	\label{fig:Lena -2 Negative}
	\end{subfigure}
	~
	\centering
	\begin{subfigure}[h]{0.14\textwidth}
	\includegraphics[width=1\textwidth]{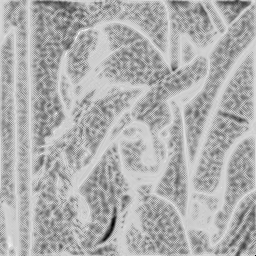}
	\caption{}
	\label{fig:Lena -3}
	
	\end{subfigure}
	~
	\centering
	\begin{subfigure}[h]{0.14\textwidth}
	\includegraphics[width=1\textwidth]{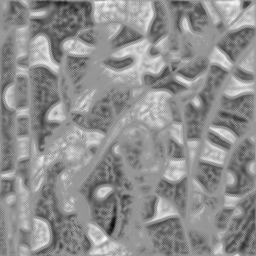}
	\caption{}
	\label{fig:Lena -3 Random}
	\end{subfigure}
	\caption{(a) Original image of size $256\times 256$. 
	(b) Poor reconstruction for $\alpha = -1$. 
	(c) Improved reconstruction for $\alpha = -2$. 
	(d) Alternative reconstruction for $\alpha = -2$. 
	(e) Poor reconstruction for $\alpha = -3$.
	(f) Alternative for $\alpha = -3$.}
	\label{fig:Recon}
\end{figure}

In this section, we demonstrate the reconstruction of an image using LFMDs and the algorithm described in Sec. \ref{sec:Locality of Polynomial Envelop and Image Reconstruction}. 
%The target image is a gray $256\times256$ Lena image. 
We use the same filters as in Sec. \ref{sec:Image Classification}, except that there is no hole at the center of the filter.
In Eq. \eqref{eq:Local Fourier-Mellin filter}, the value is set to 0 if $\sqrt{x^2 + y^2} \leq \sigma \omega_\theta $. 
For the sake of reconstruction quality, it is set to 0 only if $\sqrt{x^2 + y^2} = 0$. 
Although a hole at the center can reduce aliasing and improve classification accuracy, 
the missing information at the center results in noisy reconstructions.
Consequently, to better illustrate the idea that images can be reconstructed from LFMDs, we use descriptors with slightly decreased scale invariance.
Finally, we observe that the phase of the reconstruction depends on the initial value image and this image is random.  See Fig. \ref{fig:Lena -2}, \ref{fig:Lena -2 Negative} and Fig. \ref{fig:Lena -3}, \ref{fig:Lena -3 Random}. 

The FMT fails to reconstruct the image (Fig. \ref{fig:Lena -1}). 
In fact, reconstruction always fails when $\alpha \geq -1$.
This demonstrates that the locality of the filters is critical to reconstruction.
Best reconstruction quality (Fig. \ref{fig:Lena -2}) is achieved
when $\alpha = -2$. 
However, because the filter support is very small when $\alpha = -3$, 
it is difficult to achieve consistency in the reconstructed
local phases (Fig. \ref{fig:Lena -3 Random}).
Nevertheless, it is clear that translation invariant descriptors, such as magnitudes of Gabor wavelets \cite{Shams2002}, 
and rotation and scale invariant descriptor, such as LFMDs, are both able to reconstruct images. 
The phase information (presumed lost) is implicit in the magnitudes of the responses of the complex filters
if the filters are sufficiently well localized.

\section{Conclusion}

This paper introduced a locally rotation and scale invariant descriptor called LFMD.
The rotation and scale {\it invariance} of LFMDs computed for local patches
results in rotation, translation and scale {\it equivariance} 
of non-local image-based
representations comprised of them.
This equivariance, in turn, results in rotation, translation
and scale invariant 2D shape recognition
when combined with VLAD.
Prior work on 2D shape recognition has ignored scale, 
using image-based representations that are solely
translation and rotation equivariant \cite{Lenc2018,Weiler2017,Worrall2017,Dieleman2016,Cohen2016,Kivinen}.

We showed that a system based on LFMD-VLAD is capable of Euclidean invariant 2D shape
recognition without the necessity of training on augmented datasets.
We also showed that images can be reconstructed from LFMDs alone.
This is similar to what others have done using magnitudes 
of convolutions with Gabor wavelets. 
Results on 1) Euclidean invariant 2D shape recognition;
and 2) recovery of phases from magnitudes in images convolved with
complex filters suggest that filter locality is essential.

In future work, we plan to generalize the method described in this paper to images with cluttered backgrounds and build 
a hierarchical network exploiting the
Euclidean equivariance of non-local image-based 
representations based on LFMDs.\\\\
{\bf Acknowledgement} Xinhua Zhang gratefully acknowledges the support of the New Mexico Consortium.

{\small
\bibliographystyle{ieee}
\bibliography{wacv}
}

\end{document}